\documentclass[final,5p,times]{elsarticle}

\usepackage{prletters}
\usepackage{amsmath,graphicx}
\usepackage{times}
\usepackage{epsfig}
\usepackage{amssymb}
\usepackage{booktabs}
\usepackage{multirow}
\usepackage{subcaption}
\usepackage[export]{adjustbox}
\usepackage{float}
\usepackage{dashrule}
\usepackage{arydshln}
\usepackage[T1]{fontenc}
\usepackage[utf8]{inputenc}

\journal{Pattern Recognition Letters}

\begin{document}

\thispagestyle{empty}

\begin{frontmatter}



\title{Improving Sketch Colorization using Adversarial Segmentation Consistency}

\author[label1]{Samet Hicsonmez\corref{cor1}}
\ead{samethicsonmez@hacettepe.edu.tr}
\cortext[cor1]{Corresponding author}
\author[label2]{Nermin Samet}
\ead{nermin.samet@enpc.fr}
\author[label3]{Emre Akbas}
\ead{emre@ceng.metu.edu.tr}
\author[label1]{Pinar Duygulu}
\ead{pinar@cs.hacettepe.edu.tr}

\affiliation[label1]{organization={Hacettepe University},
            city={Ankara},
            country={Turkey}}

\affiliation[label2]{organization={LIGM, Ecole des Ponts, Univ Gustave Eiffel, CNRS},
            city={Marne-la-Vallée},
            country={France}}

\affiliation[label3]{organization={Middle East Technical University},
            city={Ankara},
            country={Turkey}}

\begin{abstract}


We propose a new method for producing color images from sketches. Current solutions in sketch colorization either necessitate additional user instruction or are restricted to the "paired" translation strategy. We leverage semantic image segmentation from a general-purpose panoptic segmentation network to generate an additional adversarial loss function. The proposed loss function is compatible with any GAN model. Our method is not restricted to datasets with segmentation labels and can be applied to unpaired translation tasks as well. Using qualitative, and quantitative analysis, and based on a user study, we demonstrate the efficacy of our method on four distinct image datasets.
On the FID metric, our model improves the baseline by up to 35 points. 
Our code, pretrained models, scripts to produce newly introduced datasets and corresponding sketch images are available at https://github.com/giddyyupp/AdvSegLoss.
\end{abstract}




\end{frontmatter}


\section{Introduction}
\label{sec:intro}

The task of image generation from an input sketch or edge map is known as "sketch to image translation", or "sketch colorization". Sketches capture essential content of the images and they can be easily acquired. Yet, vast amount of domain difference between single channel edge maps and color images makes sketch colorization a challenging process. Lack of details in sketches especially for  background is another problem. 

Sketch colorization has been investigated in numerous domains: faces~\cite{deepfacedrawing,lee2020reference,linestofacephoto,multi_density}, objects~\cite{sketchygan,texturegan,lu2018image,liuunsupervised}, animes~\cite{comicolorization,ci2018user,autopainter,zhang2018two,zhang2017style,Yuan_2021_CVPR,tip_dual_color}, art~\cite{liu2020sketch}, icons~\cite{Li_2022_CVPR} and scenes~\cite{scribbler,edgegan,zou2019language}. The majority of the approaches require user direction in the form of supplementary input, such as a reference color, patch, or image. These approaches usually generate surreal colorizations otherwise. Except for a few studies (e.g., Liu et al.~\cite{liuunsupervised}), most of the approaches follow the ``paired'' strategy, which is restricted to use datasets with a ground-truth image for each sketch. 

In this study, we aim to leverage general purpose semantic image segmentation to alleviate the  aforementioned shortcomings.
We argue that an accurately colored sketch would produce a ``real'' segmentation result, i.e., a result that looks like the segmentation of a real image. Thus, for sketch based image colorization problem, we exploit semantic segmentation methods that have reached to a degree of maturity even for datasets on which they were not trained  (Section~\ref{sec:dataset}). We introduce a segmentation-based adversarial loss to be used in a GAN (Generative Adversarial Network) setup. With our approach, neither extra user instruction nor "paired" input is required.








\begin{figure*}[h]
\centering
\includegraphics[width=0.90\textwidth]{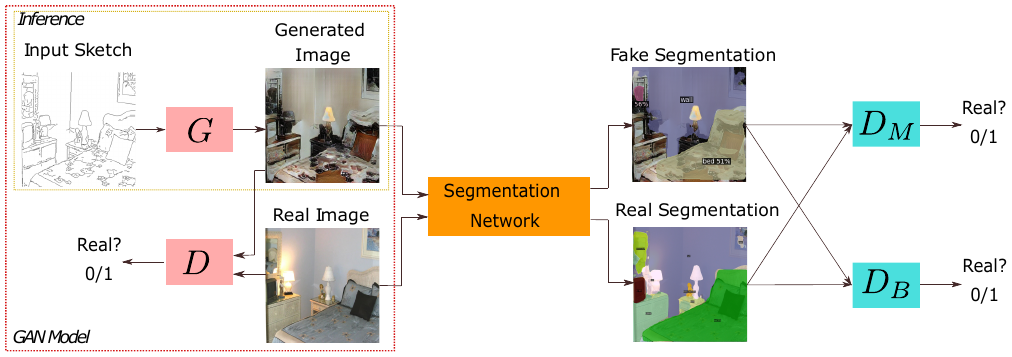}
\caption{The proposed model for sketch colorization with Adversarial Segmentation Loss (ASL). It is composed of two parts; a general purpose image translation GAN model, and an image segmentation model. During training, input sketches are first colorized using the baseline GAN model. Then, generated and ground truth color images are fed to the pre-trained panoptic segmentation model to extract fake and real segmentation maps. Finally, two additional discriminators are used to classify the segmentation maps as fake or real, respectively. The box with dashed yellow borders shows the inference stage. Red border marks the GAN model used for sketch to image translation. Here, Pix2Pix is used as an example image translation model, which could be replaced by any paired or unpaired model.}
\label{fig:model}
\end{figure*}


We introduce three models for varying levels of segmentation feedback in the sketch to image translation pipeline. Our models could be integrated into both paired and unpaired GAN models. 
We illustrate the effectiveness of applying segmentation cues via comprehensive experimental analyses. 
This paper extends our previous work~\cite{samhi_icip} in the following ways: (i) We apply our method to a new task: label-to-photo translation. Our experiments on two challenging datasets show that our segmentation-based adversarial loss is useful in this task, too. Again, ground-truth segmentation labels are not a requirement for our approach.
(ii) We perform experiments to set the optimal values for weights of two additional discriminators. (iii) We incorporate an outdoor dataset, Cityscapes~\cite{Cordts2016Cityscapes}, to both sketch colorization and label to photo translation tasks. (iv) We add a new metric, mean Intersection over Union (mIoU), in addition to FID score to measure the performance of all the models more reliably.



\section{Related Work}
\label{sec:related_work}


Even though the sketch and the edge map of an image are different concepts, in practice, XDoG~\cite{xdog} or HED~\cite{hed} based edge maps are considered as sketches (e.g.,~\cite{scribbler, autopainter}). Moreover, some sketch based models~\cite{sketchygan} use edge maps for data augmentation. Hence, we refer to all these models as "sketch-to-image translation" or "sketch colorization" models. Although general purpose image-to-image translation methods~\cite{cyclegan, dualgan, pix2pix, ganilla, huang2018multimodal} could be used for sketch-to-image translation tasks, the results are not satisfactory.

One widely used solution to improve the colorization performance is to employ additional  color~\cite{scribbler,autopainter,zhang2018two,ci2018user,Yuan_2021_CVPR,tip_dual_color}, patch~\cite{texturegan}, image~\cite{lee2020reference,zhang2017style,liu2020sketch,comicolorization,Li_2022_CVPR,Lee_2020_CVPR} or language guidance~\cite{zou2019language,Zou_2018_ECCV,Kim_2019_ICCV}. For instance, in color guidance, users specify their desired colors for the regions in the sketch image, and the model utilizes this information to generate the same or similar colors for these regions. Some automatic methods also utilize user guidance to improve their performance as a hybrid approach. Most of the sketch-to-image translation methods are based on ``paired’’ training approach~\cite{scribbler,autopainter,sketchygan,edgegan}, but, recently unpaired methods have also been presented~\cite{liuunsupervised,liu2020sketch}.


Scribbler~\cite{scribbler} presents one of the very first paired and user guided scene sketch colorization models. In addition to pixel, perceptual and GAN losses, Scribbler uses total variation loss to encourage smoothness. XDoG is used to generate sketch images of 200k bedroom photos.
DCSGAN~\cite{tip_dual_color} uses HSV color space in addition to the RGB, for line art colorization task. 
Zou et al.~\cite{zou2019language} use text inputs to progressively colorize an input sketch, in such a way that a novel text guided sketch segmenter locates the objects in the scene.
EdgeGAN~\cite{edgegan} maps edge images to a latent space during training using an edge encoder. During inference, the edge encoder is used to encode the input sketch to the latent space to subsequently generate a color image. Experiments are provided for 14 foreground and 3 background objects from COCO~\cite{coco} dataset.

EdgeGAN~\cite{edgegan} and Scribbler~\cite{scribbler} use a supervised approach where input sketches and corresponding output images exist. However, it is hard to collect sketch image pairs.
Liu et al.~\cite{liuunsupervised} propose a two stage method to convert object sketches to color images in an unsupervised (unpaired) way. They first convert sketches to gray scale images, and then to color images. Self supervision is used to complete the deliberately deleted sketch parts and clear the added noisy edges from sketch images. 

In Sketch-to-Art~\cite{liu2020sketch}, an art image is generated using an input sketch, with the additional help of the target style art image. Content of the input sketch and style of the art image are encoded, and then fused to generate a stylized art image. In~\cite{Li_2022_CVPR} authors proposed a method to colorize icons which utilizes colored icon images as input in addition to the black-white icons. 

The user input is valuable not only in helping the colorization network to put the right colors to indicated regions, but also in removing the color bleeding problems. In~\cite{Kim_2021_ICCV}, users draw scribbles to the regions on the generated image suffered from color bleeding artifacts for guiding the model to fix them. 

Unlike these methods, our method does not require any user input to generate satisfactory colorization. Instead, we utilize adversarial segmentation guidance to improve performance.


\section{Adversarial Segmentation Consistency}
\label{sec:advsegloss_model}

Figure~\ref{fig:model} shows the overall structure of our proposed model for sketch colorization, which we refer to as Adversarial Segmentation Loss (ASL) based model. In this work, we used \textit{Pix2Pix} and \textit{CycleGAN} methods as our baselines for paired and unpaired training, respectively. This preference is made based on the effectiveness of these methods across a variety of tasks and datasets. In the figure, Pix2Pix is used to show ASL based model for paired approach. Our model could be integrated into any other paired or unpaired GAN model.

Our model consists of a baseline GAN, a panoptic segmentation network (\textit{Seg}) and two discriminators (\textit{$D_{M}$} and \textit{$D_{B}$}). Panoptic segmentation network is trained offline on the COCO Stuff~\cite{coco_stuff} dataset and 
its weights are frozen during the training of our model. Fake and real images are fed to the \textit{Seg} network to get real and fake segmentation maps. Then, these two segmentation maps are given to the discriminators to classify them as fake or real. We designed three variants of our model to embed different levels of segmentation feedback to the sketch to image translation pipeline.


The first variant utilizes the full (multiclass) segmentation map of an image where all foreground and background classes (a total of 135 classes) are considered. In this model, ground-truth color image $I_{real}$ and the generated color image $I_{fake}$ are fed to \textit{Seg} which outputs full segmentation maps for both images. Then, these two outputs are given to a discriminator network \textit{$D_{M}$} to discriminate between real and fake segmentation maps. We call this model as \textit{Multi-class} in the rest of the paper.

As a higher level of abstraction, grouping objects as background and foreground alone may yield sufficient information. The second variant of our model 
uses only two classes (background and foreground) in the segmentation map by grouping all foreground classes into one and all background classes into another class. 
In this model, which we refer to as \textit{Binary},
binary segmentation outputs for real and fake images are fed to a discriminator network \textit{$D_{B}$} to discriminate between real and fake ones. Finally, our third variant is the union of the above two. It contains both discriminators, and is named as \textit{Combined}. 

Overall loss function for our model is the sum of losses of the baseline GAN model ($L_G$) and the two additional discriminators’ ($L_B$ and $L_M$). 
That is, the objective function is:
\begin{equation*}
\begin{aligned}
 \label{eq:loss}
 \mathcal{L} = w_gL_G + w_bL_B + w_mL_M
  \end{aligned}
\end{equation*}

Let $Seg_B$ and $Seg_M$ correspond to the panoptic segmentation networks in \textit{Binary} and \textit{Multi-class} cases. The additional losses that we introduce, ${L}_{B}$ and ${L}_{M}$, are defined as:

\begin{equation*}
\begin{aligned}
 \label{eq:adv_loss_b}
  \mathcal{L}_{B}(G, D_B, Seg_B) = {} &  
  \sum_{i}log(D_B(Seg_B(y_i))) +  \\
  & \sum_{i}log(1 - D_B(Seg_B(G(x_i)))
  \end{aligned}
\end{equation*}
\begin{equation*}
\begin{aligned}
 \label{eq:adv_loss_m}
  \mathcal{L}_{M}(G, D_M, Seg_M) = {} &  
  \sum_{i}log(D_M(Seg_M(y_i))) +  \\
  & \sum_{i}log(1 - D_M(Seg_M(G(x_i)))
  \end{aligned}
\end{equation*}

Let $x_i$ be an input sketch image, and $y_i$ be the corresponding ground truth color image. When the baseline GAN model is Pix2Pix~\cite{pix2pix}, GAN loss $L_G$ is formulated as:
\begin{equation*}
 \label{eq:adv_loss_pix2pix}
   \mathcal{L}_{G}(G, D) = {}   
  \sum_{i}log(D(x_i)) + \sum_{i}log(1 - D(G(x_i)) + \sum_{i} \| y_i - G(x_i)\|
\end{equation*}
When baseline is CycleGAN~\cite{cyclegan}, $L_G$ for direction $X \rightarrow Y$  is:
\begin{equation*}
 \label{eq:adv_loss_cyclegan}
 \begin{aligned}
   \mathcal{L}_{G}(G_X, G_Y, D_X) = {}  & 
  \sum_{j}log(D_X(y_j)) +  \sum_{i}log(1 - D_X(G_X(x_i)) \\ & +   \sum_{i} \| x_i - G_Y(G_X(x_i))\|
  \end{aligned}
\end{equation*}
where $G_X$ maps input sketches to color images, and $G_Y$ maps the color images back to sketch domain. $D_X$ is the discriminator for domain $X$, i.e. sketches. Final $L_G$ for CycleGAN is the sum of above formulation for two directions.

We analysed the effect of each component in the objective function and, set $w_g$, $w_b$ and $w_m$ to $1$ based on the experimental analysis (see Section~\ref{sec:quant_analysis}). Note that, for the \textit{Binary} model  $w_m$, and for the \textit{Multi-class} model $w_b$ is set to 0 respectively.

\begin{figure}
\captionsetup[subfigure]{labelformat=empty}
\centering
\begin{subfigure}[b]{0.15\textwidth}
\includegraphics[width=1.0\textwidth]{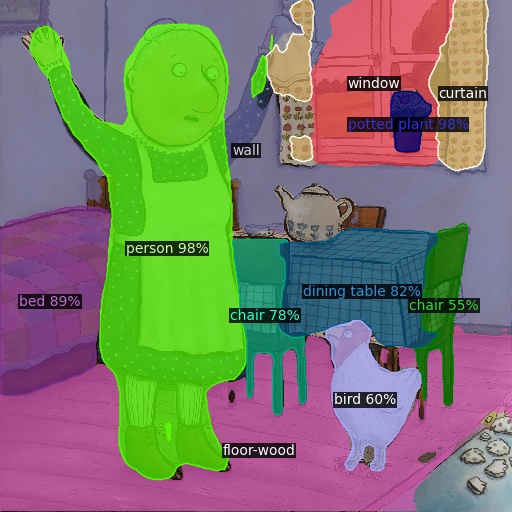}
\end{subfigure}
\begin{subfigure}[b]{0.15\textwidth}
\includegraphics[width=1.0\textwidth]{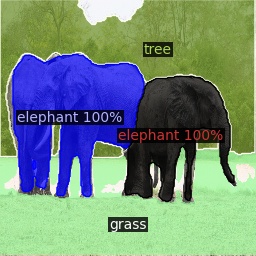}
\end{subfigure}
\begin{subfigure}[b]{0.15\textwidth}
\includegraphics[width=1.0\textwidth]{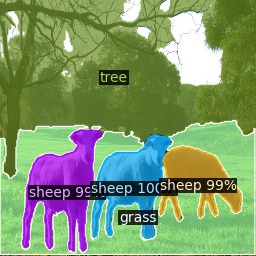}
\end{subfigure}
\begin{subfigure}[b]{0.22\textwidth}
\includegraphics[width=1.0\textwidth]{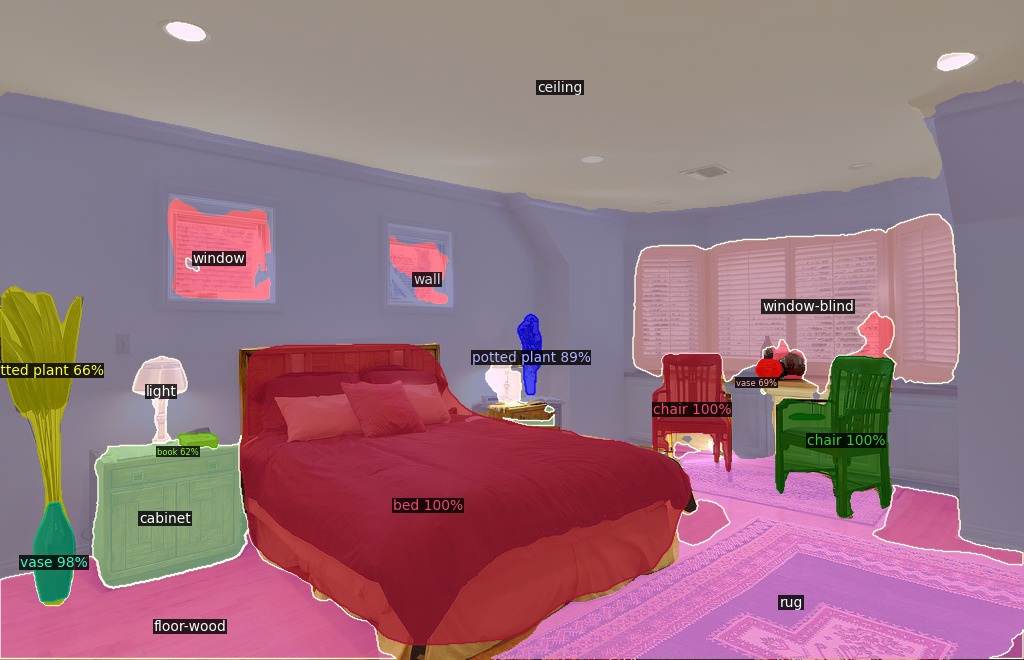}
\end{subfigure}
\begin{subfigure}[b]{0.22\textwidth}
\includegraphics[width=1.0\textwidth]{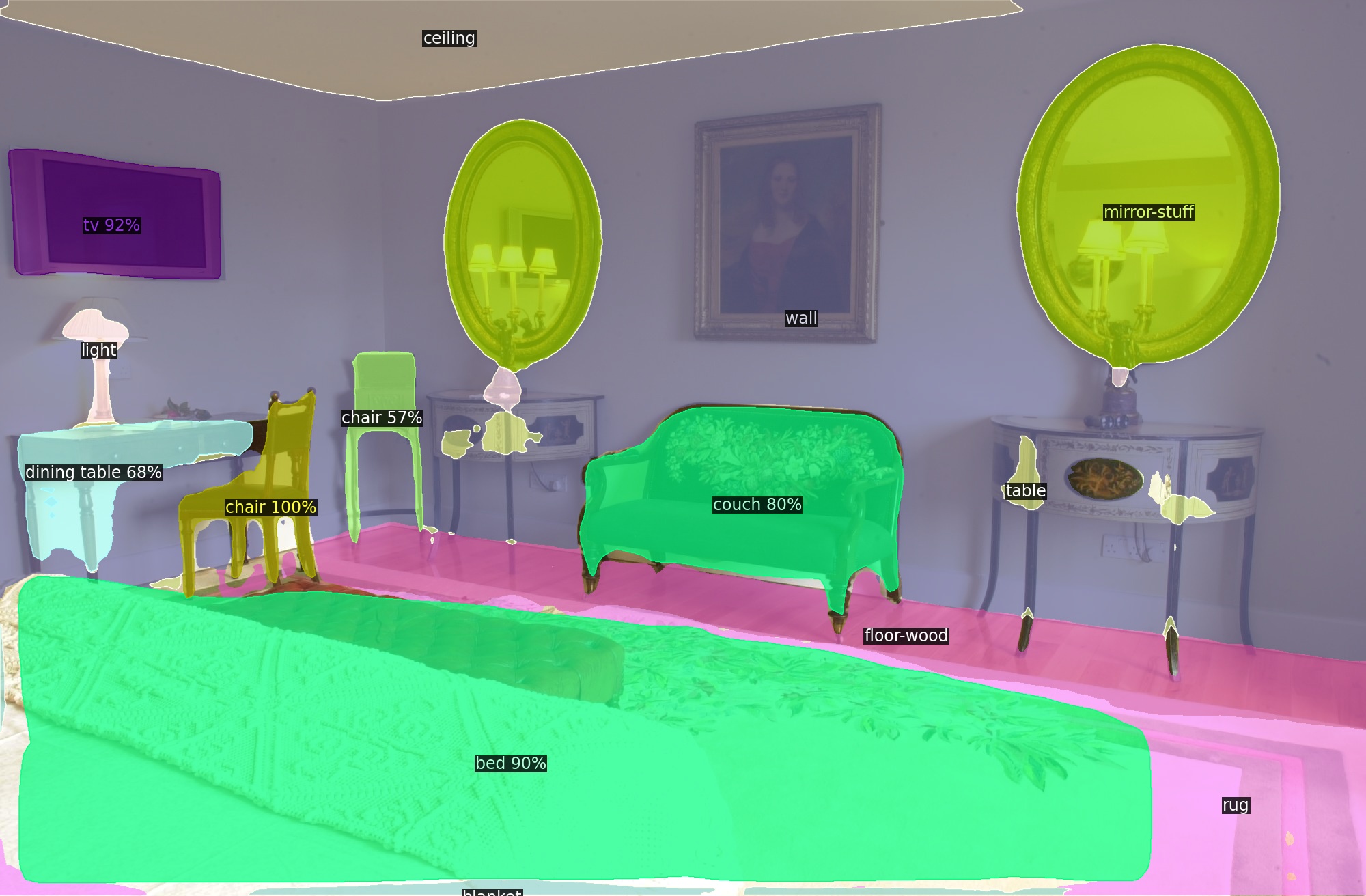}
\end{subfigure}
\caption{Sample segmentations using general purpose panoptic segmentation network on different datasets. The model generalizes well to several domains.}
\label{fig:seg-samples}
\end{figure}

\begin{figure*}[ht]
\centering
\begin{tabular}{cccccc}
\shortstack{Input \\ { }} & \shortstack{GT \\ { }} & \shortstack{CycleGAN \\ { }}  & \shortstack{C-ASL \\ (Multi-class)} & \shortstack{C-ASL\\ (Binary)} & \shortstack{C-ASL\\(Combined)} \\
\includegraphics[width=0.14\textwidth,  ,valign=m, keepaspectratio,]{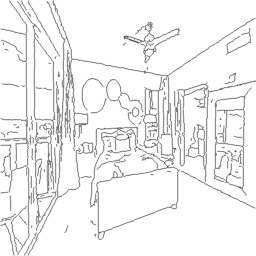}  \hspace*{-12pt} &
\includegraphics[width=0.14\textwidth,  ,valign=m, keepaspectratio,]{./figures/visual_res/ade20k/unpaired/gt/ADE_val_00000129.jpg} \hspace*{-12pt} &
\includegraphics[width=0.14\textwidth,  ,valign=m, keepaspectratio,]{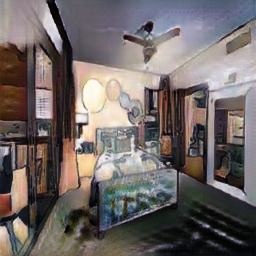} \hspace*{-12pt} &
\includegraphics[width=0.14\textwidth,   ,valign=m, keepaspectratio,]{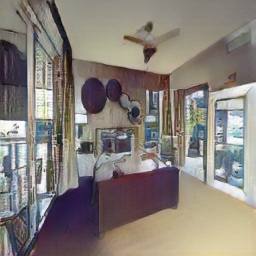} \hspace*{-12pt} &
\includegraphics[width=0.14\textwidth,   ,valign=m, keepaspectratio,]{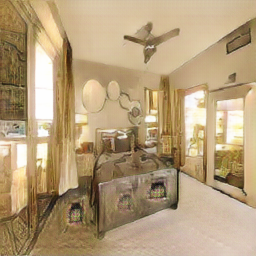} \hspace*{-12pt} &
\includegraphics[width=0.14\textwidth,   ,valign=m, keepaspectratio,]{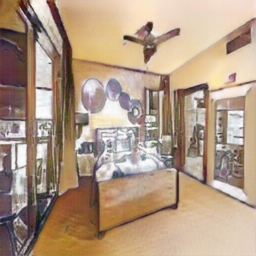} \hspace*{-12pt} \\
\noalign{\vskip 1mm}
\includegraphics[width=0.14\textwidth,  ,valign=m, keepaspectratio,]{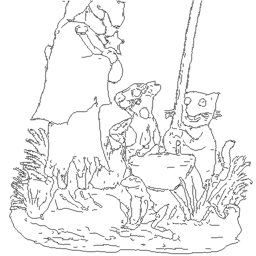}  \hspace*{-12pt}&
\includegraphics[width=0.14\textwidth,  ,valign=m, keepaspectratio,]{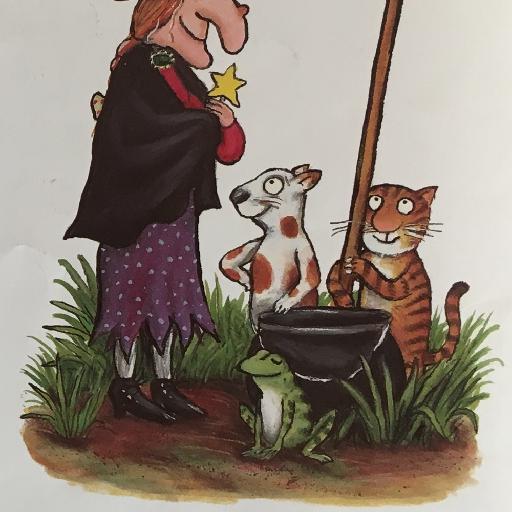} \hspace*{-12pt}&
\includegraphics[width=0.14\textwidth,  ,valign=m, keepaspectratio,]{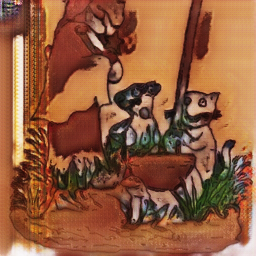} \hspace*{-12pt}&
\includegraphics[width=0.14\textwidth,   ,valign=m, keepaspectratio,]{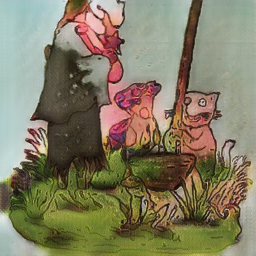} \hspace*{-12pt} &
\includegraphics[width=0.14\textwidth,   ,valign=m, keepaspectratio,]{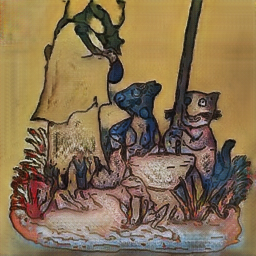} \hspace*{-12pt} &
\includegraphics[width=0.14\textwidth,   ,valign=m, keepaspectratio,]{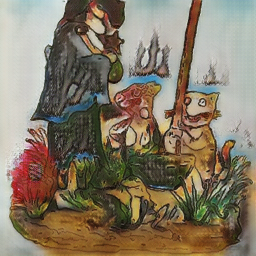} \hspace*{-12pt}\\
\midrule
\shortstack{Input \\ { }} & \shortstack{GT \\ { }} & \shortstack{Pix2Pix \\ { }}  & \shortstack{P-ASL \\ (Multi-class)} & \shortstack{P-ASL\\ (Binary)} & \shortstack{P-ASL\\(Combined)} \\
\includegraphics[width=0.14\textwidth,  ,valign=m, keepaspectratio,]{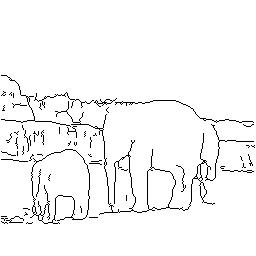}  \hspace*{-12pt}&
\includegraphics[width=0.14\textwidth,  ,valign=m, keepaspectratio,]{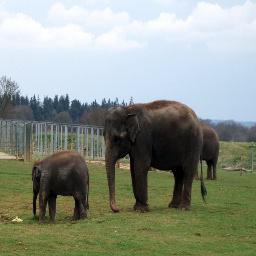} \hspace*{-12pt}&
\includegraphics[width=0.14\textwidth,  ,valign=m, keepaspectratio,]{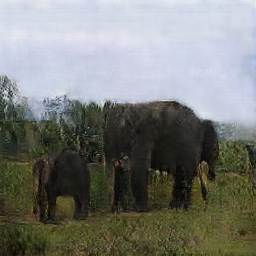} \hspace*{-12pt}&
\includegraphics[width=0.14\textwidth,   ,valign=m, keepaspectratio,]{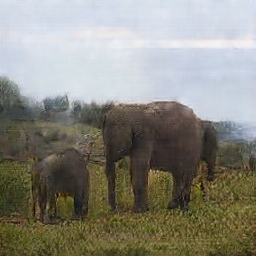} \hspace*{-12pt} &
\includegraphics[width=0.14\textwidth,   ,valign=m, keepaspectratio,]{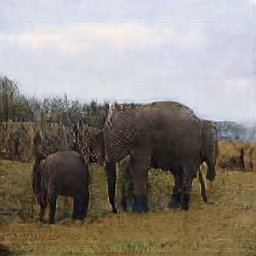} \hspace*{-12pt} &
\includegraphics[width=0.14\textwidth,   ,valign=m, keepaspectratio,]{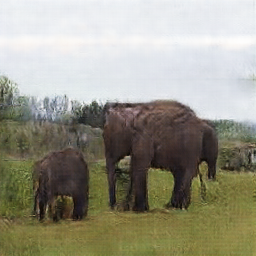} \hspace*{-12pt}\\
\noalign{\vskip 1mm}
\includegraphics[width=0.14\textwidth,  ,valign=m, keepaspectratio,]{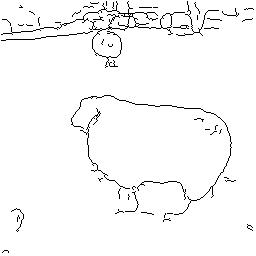}  \hspace*{-12pt}&
\includegraphics[width=0.14\textwidth,  ,valign=m, keepaspectratio,]{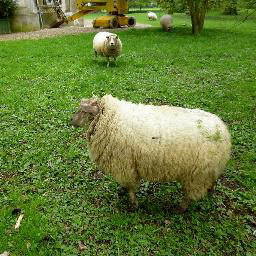} \hspace*{-12pt}&
\includegraphics[width=0.14\textwidth,  ,valign=m, keepaspectratio,]{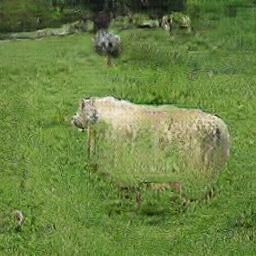} \hspace*{-12pt}&
\includegraphics[width=0.14\textwidth,   ,valign=m, keepaspectratio,]{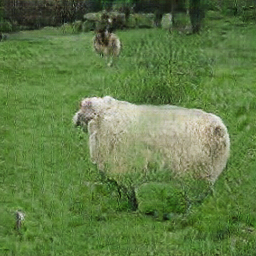} \hspace*{-12pt} &
\includegraphics[width=0.14\textwidth,   ,valign=m, keepaspectratio,]{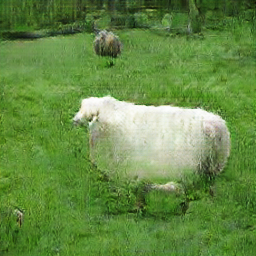} \hspace*{-12pt} &
\includegraphics[width=0.14\textwidth,   ,valign=m, keepaspectratio,]{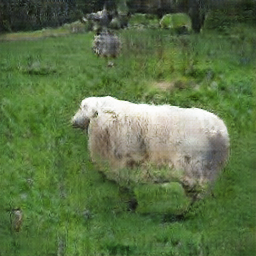} \hspace*{-12pt}  \\
\end{tabular}
\caption{Sample results from baselines and our model with different settings. Input images on each row are from bedroom, illustration, elephants and sheep datasets, respectively. First two rows display results of unpaired training (baseline is CycleGAN), and last two rows show results for paired training (baseline is Pix2Pix).
On bedroom and elephant datasets \textit{Binary}, on illustration and sheep datasets \textit{Combined} setting gave best results for both training schemes.}
\label{fig:all-results}
\end{figure*}

\section{Datasets}
\label{sec:dataset}

We evaluated our models on five challenging datasets (see Table~\ref{tab:advsegloss_dataset}). The first dataset consists of bedroom images from the ADE20k indoor dataset~\cite{ade20k}, with $1355$ train and $135$ test images. The second dataset is Cityscapes~\cite{Cordts2016Cityscapes} dataset which contains $2975$ training and $500$ test images. The third dataset~\cite{ganilla} contains illustrations from children’s books by Alex Scheffler, with $659$ train and $131$ test images. The fourth and fifth ones were curated by us from the COCO dataset. We collected images containing elephant or sheep. Note that these images may also contain other foreground/background objects such as person, animals, mountains, grass and sky. Elephant dataset contains $1800$ train and $343$ test images, and the sheep dataset has $1300$ train and $229$ test images. 
Example images from these datasets and their segmentation outputs are shown in Figure~\ref{fig:seg-samples}.


\begin{table}
\caption{Statistics of the datasets used in our experiments.}
\label{tab:advsegloss_dataset}
\centering
\resizebox{0.75\columnwidth}{!}{
\begin{tabular}{lcc}
\hline
Dataset & Train Images &  Test Images \\
\hline
ADE20k Bedroom       & 1355    &  135   \\
Cityscapes          & 2975 & 500 \\
Illustration & 659    &  131     \\
COCO Elephant       & 1800    &  343     \\
COCO Sheep             & 1300    & 229     \\ 
\hline
\end{tabular}}
\end{table}

Edge images are extracted using the HED~\cite{hed} method. In the first two columns of Figure~\ref{fig:all-results}, we present sample natural and edge images for all the datasets. It can be seen that the images contain a variety of foreground and background objects, also it is hard to figure out the source dataset for some images.

\begin{table}
\caption{Comparison with CycleGAN~\cite{cyclegan} on \textbf{unpaired sketch-to-image translation} task in terms of \textbf{FID scores}, lower is better.}
\label{tab:advsegloss_fid_scores_unpaired_s2p}
\begin{center}
\resizebox{\columnwidth}{!}{
\begin{tabular}{ccccc}
 \toprule 
\shortstack{Dataset\\ { }} & \shortstack{CycleGAN\\ { }} & \shortstack{C-ASL\\(Multi-class)} & \shortstack{C-ASL\\(Binary)} & \shortstack{C-ASL\\(Combined)} \\
 \midrule 
Bedroom         & 113.1 & 111.7 & \textbf{87.1} & 93.2       \\
Cityscapes         & 62.9 & 64.1 & 64.9 & \textbf{59.1}          \\
Illustration         & 213.6 & 206.9 & 204.8 & \textbf{189.4}          \\
 Elephant        & 126.4 & 103.9 & \textbf{91.9} & 116.9      \\
Sheep                & 209.3 & 207.2 & 236.1 & \textbf{196.8}     \\
 \bottomrule 
\end{tabular}}
\end{center}
\vspace{-5mm}
\end{table}

\section{Experiments}
\label{sec:exp}

We used PyTorch~\cite{pytorch} to implement our models.
We use sketch images as source domain, and color images as target domain. All training images (i.e. color and sketch images) are resized to $256\times256$ pixels. We train all models for $200$ epochs using the Adam optimizer~\cite{adam} with a learning rate of $0.0002$. We conducted all our experiments on a NVIDIA Tesla V100 GPU.

We compared our models with Pix2Pix~\cite{pix2pix} and AutoPainter (AP)~\cite{autopainter} for paired and CycleGAN~\cite{cyclegan} for unpaired setting on \textbf{sketch-to-image translation} task. We used the official implementations that are publicly available. Baseline models are trained for $200$ epochs. 
Our proposed ASL model that uses Pix2Pix as the baseline GAN model is referred to as P-ASL, and similarly C-ASL refers to the model that uses CycleGAN.

\subsection{Quantitative Analysis}
\label{sec:quant_analysis}

To quantitatively evaluate the quality of generated images, we used the widely adopted Frechet Inception Distance (FID)~\cite{FID} metric. FID score measures the distance between the distributions of the generated and real images. Lower FID score indicates the higher similarity between two image sets.

\begin{table}
\caption{Comparison with AutoPainter~\cite{autopainter} and Pix2Pix~\cite{pix2pix} on \textbf{paired sketch-to-image translation} task in terms of \textbf{FID scores}, lower is better.}
\label{tab:advsegloss_fid_scores_paired_s2p}
\begin{center}
\resizebox{\columnwidth}{!}{
\begin{tabular}{cccccc}
 \toprule 
\shortstack{Dataset\\ { }} & \shortstack{Auto\\{Painter}} & \shortstack{Pix2Pix\\ { }} & \shortstack{P-ASL\\(Multi-class)} & \shortstack{P-ASL\\(Binary)} & \shortstack{P-ASL\\(Combined)} \\
 \midrule 
Bedroom         & 206.8 & 100.5 & 100.0 & \textbf{95.1} & 110.1      \\
Cityscapes      & 151.3  & 74.1 & \textbf{69.9} & 71.6  & 71.2          \\
Illustration    & 272.0 &180.0 & 176.9 & 178.0 & \textbf{175.7}          \\
Elephant        & 155.1 & 83.5   & 85.8   & \textbf{78.8} & 82.8      \\
Sheep           & 233.1 & 157.0 & 159.9 & 162.0 & \textbf{150.5}     \\
 \bottomrule 
\end{tabular}}
\end{center}
\vspace{-5mm}
\end{table}

On Bedroom and Cityscapes datasets where ground truth segmentation maps are available, we also calculate the mean Intersection over Union (mIoU) scores on colorized images. We forward each colorized image to an off-the-shelf segmentation model trained on these two datasets separately. mIoU score measures the quality of the segmentation. We argue that better colorized images should yield higher mIoU scores.

We present FID scores for unpaired translation in Table~\ref{tab:advsegloss_fid_scores_unpaired_s2p} and  paired translation in Table~\ref{tab:advsegloss_fid_scores_paired_s2p}. FID scores are inline with the visual inspections (see Figure~\ref{fig:all-results}), for all the datasets, at least one variant of our model performed better than the baseline.

First of all, when we compare FID scores of two training schemes and baseline models, paired training (Pix2Pix) performed better than unpaired training, as expected. However, our ``adversarial segmentation loss’’ affected the results of paired and unpaired cases differently. For instance, on elephant dataset our models improved baseline up to $35$ points for unpaired case, but only $5$ points for paired case.

Another crucial observation is that segmentation guidance closed the gap between unpaired and paired training results. Best FID scores for unpaired models on bedroom, illustration and elephant datasets become very close to or even better than paired training. For instance on the elephant dataset, the initial $40+$ point FID gap ($126$ vs $83$) dropped to $13$ ($92$ vs $79$) on \textit{Binary} setting. Here the only exception is the sheep dataset. Since the sheep dataset contains various complex objects, unpaired and paired models failed to generate plausible images.

\begin{table}
\caption{Comparison with CycleGAN on \textbf{unpaired sketch-to-image translation} task in terms of \textbf{mIoU scores}, higher is better.}
\label{tab:advsegloss_miou_scores_unpaired_s2p}
\begin{center}
\resizebox{\columnwidth}{!}{
\begin{tabular}{cccccc}
 \toprule 
\shortstack{Dataset\\ { }} & \shortstack{CycleGAN\\ { }}  & \shortstack{C-ASL\\(Multi-class)} & \shortstack{C-ASL\\(Binary)} & \shortstack{C-ASL\\(Combined)} & \shortstack{Oracle \\ { }}\\
 \midrule 
Bedroom         & 5.20 &   \textbf{6.71} & 6.58 & 6.44 &  20.62    \\
Cityscapes      & 15.67 &  14.63 & 13.56 & \textbf{15.68}  &  44.85      \\
 \bottomrule 
\end{tabular}}
\end{center}
\vspace{-5mm}
\end{table}

\begin{table}
\caption{Comparison with Pix2Pix on \textbf{paired sketch-to-image translation} task in terms of \textbf{mIoU scores}, higher is better.}
\label{tab:advsegloss_miou_scores_paired_s2p}
\begin{center}
\resizebox{\columnwidth}{!}{
\begin{tabular}{ccccccc}
 \toprule 
\shortstack{Dataset\\ { }} & \shortstack{Auto\\ {Painter}}  &
\shortstack{Pix2Pix\\ { }}  & \shortstack{P-ASL\\(Multi-class)} & \shortstack{P-ASL\\(Binary)} & \shortstack{P-ASL\\(Combined)} & \shortstack{Oracle \\ { }}\\
 \midrule 
Bedroom         & 2.08 & 6.49 & 6.95  &  \textbf{7.39} & 6.60 &  20.62    \\
Cityscapes      & 6.02  & 18.71 & 18.63 & 18.70  & \textbf{18.74}  &  44.85      \\
 \bottomrule 
\end{tabular}}
\end{center}
\vspace{-5mm}
\end{table}

\begin{table}
\caption{User Study results.}
\label{tab:user_study}
\begin{center}
\begin{tabular}{lcc}
\toprule 
 Dataset & CycleGAN & C-ASL \\
 \midrule 
Bedroom         & 20.0 & \textbf{80.0}  \\
Illustration        & 27.0 & \textbf{73.0}  \\
Elephant        & 39.1 & \textbf{60.9}  \\
Sheep                & 19.1 & \textbf{80.9}  \\
\hline
Cityscapes (L2P)         & 25.7  & \textbf{74.3}  \\
 \bottomrule 
\end{tabular}
\end{center}
\vspace{-5mm}
\end{table}

\begin{table}
\caption{Effect of changing the $w_b$ and $w_m$ values, $w_b$ is used as $1.0$ for all experiments. Using $1.0$ for both weights yields the best FID score on the ADE20k bedroom images for the task of sketch-to-image translation.}
\label{tab:advsegloss_abl}
\centering
\resizebox{0.70\columnwidth}{!}{
\begin{tabular}{lccccc}
\toprule 
& \multicolumn{5}{c}{$w_b$ and $w_m$}   \\
\cmidrule(lr){2-6}
& 0.1 & 0.5 & 1.0 & 5.0 & 10.0 \\
\midrule
FID & 114.8 & 114.5 & \textbf{93.2} & 147.8 & 104.6 \\
 \bottomrule 
\end{tabular}} 
\vspace{-3mm}
\end{table}

We show mIoU scores for unpaired translation in Table~\ref{tab:advsegloss_miou_scores_unpaired_s2p} and paired translation in Table~\ref{tab:advsegloss_miou_scores_paired_s2p}. We also present oracle performances of the segmentation method on both datasets. On mIoU metric, again for all the datasets, at least one of the variants of our model performed better than the baseline.

When we look at the best performing settings on different datasets, structure of the dataset has an effect on the results. For instance, even though one is an indoor and the other one is an outdoor dataset, bedroom and elephant images are composed of similar structure. FG/BG ratios and placements of them in these datasets are similar across all images, i.e. walls, ceiling and floors in bedroom images are always positioned in the same places on different images. Also elephant images contain very few FG objects, i.e. only elephants most of the time, and large BG areas such as grass, trees and sky. On these two datasets, \textit{Binary} setting which considers FG/BG classes only gave the best FID score.
On the other hand, illustration and sheep images got a variety of FG objects and scenes. On such datasets, using only a FG/BG discriminator even degrades the performance.


Our model has two important parameters, $w_b$ and $w_m$, to control the effect of segmentation discriminators. To find the best possible values, we conducted experiments by training our models on the ADE20k bedroom images on the unpaired sketch to image translation task (see Table~\ref{tab:advsegloss_abl}). Using a small value like $0.1$ gives a similar score to baseline CycleGAN. On the other hand using a big value like $5.0$ increased the FID score dramatically. Setting $w_b$ and $w_m$ to $1.0$ resulted in the best FID score, thus the weights are set to $1.0$ in all experiments.

\subsection{Qualitative Analysis and User Study}

We present visual results of sketch colorization for our model and the baseline models in the Figure~\ref{fig:all-results}.
On bedroom and illustration datasets, we show results of unpaired training, and on elephant and sheep datasets we show paired training results.

On the bedroom dataset, the \textit{Binary} setting generates better images compared to baselines and other settings. Colors are uniform across the object parts in this setting. There are defective colors in the CycleGAN results such as the bottom of the bed and floor. On the illustration dataset, the baseline model performed poorly. Objects are hard to recognize and most importantly colors are not proper at all. On the other hand, \textit{Multi-class} and \textit{Combined} settings generate significantly better images i.e. generated objects and background got consistent colors. Finally, on elephant and sheep datasets although generated images are not very visually appealing for all the methods, segmentation guided images are quite appealing compared to baseline models’. On the elephant dataset \textit{Binary}, on the sheep dataset \textit{Combined} setting performed the best.

We conducted a user study to measure realism of generated images. We show two random images (at random positions, left or right) which were generated with CycleGAN and our best setting (lowest FID score) for all four datasets, and asked participants to select the more realistic one. 

We collected a total of 115 survey inputs from 39 different users. We ask users to evaluate 4 S2P models and 1 L2P model. In Table~\ref{tab:user_study}, we present results of the user study in terms of preference percentages of each model.  User study results are inline with the FID score results, on all datasets, images generated by our model were preferred by the users most of the time.
On sheep and elephant datasets, users struggled to select an answer. Color distributions and shapes of FG objects are two dominant factors which lead user preferences.

\begin{figure}
\captionsetup[subfigure]{labelformat=empty}
\centering
\begin{subfigure}[b]{0.11\textwidth}
 \caption{Input}
\includegraphics[width=1.0\textwidth]{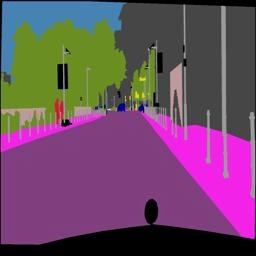}
 \vspace{0.05 cm}
\includegraphics[width=1.0\textwidth]{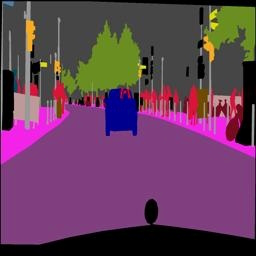}
 \vspace{0.05 cm}
\includegraphics[width=1.0\textwidth]{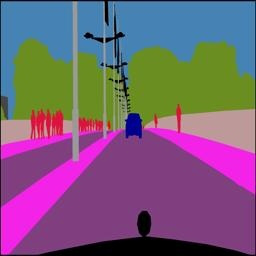}
\end{subfigure}
\begin{subfigure}[b]{0.11\textwidth}
 \caption{GT}
\includegraphics[width=1.0\textwidth]{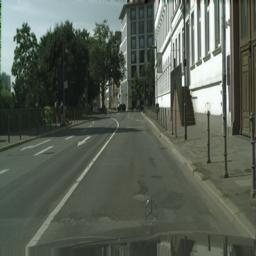}
 \vspace{0.05 cm}
\includegraphics[width=1.0\textwidth]{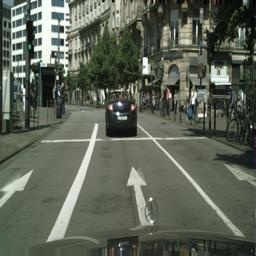}
 \vspace{0.05 cm}
\includegraphics[width=1.0\textwidth]{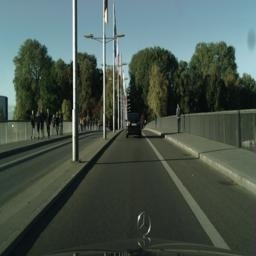}
\end{subfigure}
\begin{subfigure}[b]{0.11\textwidth}
 \caption{CycleGAN}
\includegraphics[width=1.0\textwidth]{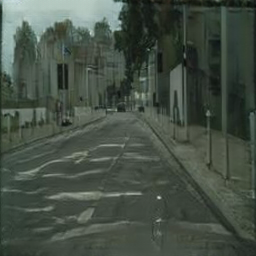}
 \vspace{0.05 cm}
\includegraphics[width=1.0\textwidth]{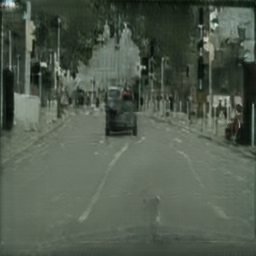}
 \vspace{0.05 cm}
\includegraphics[width=1.0\textwidth]{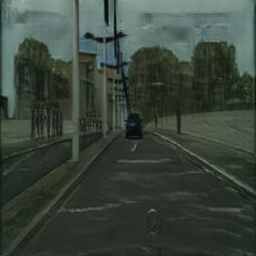}
\end{subfigure}
\begin{subfigure}[b]{0.11\textwidth}
 \caption{Binary}
\includegraphics[width=1.0\textwidth]{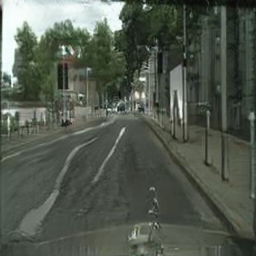}
 \vspace{0.05 cm}
\includegraphics[width=1.0\textwidth]{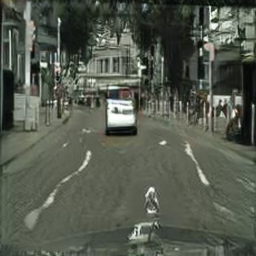}
 \vspace{0.05 cm}
\includegraphics[width=1.0\textwidth]{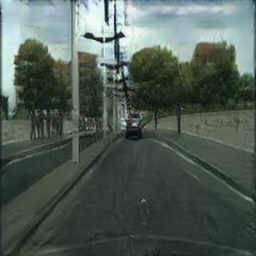}
\end{subfigure}
\caption{Sample results on Cityscapes dataset for CycleGAN and our best model. In the first image, CycleGAN generates buildings instead of trees, also there are defects on the road. Other two images are blurry and lack details. On the other hand, in all cases our \textit{Binary} setting generates more visually appealing images.}
\label{fig:city-results}
\end{figure}

\begin{table}
\caption{Comparison with CycleGAN~\cite{cyclegan} on \textbf{unpaired label-to-photo translation}  task in terms of \textbf{FID scores}, lower is better.}
\label{tab:advsegloss_fid_scores_unpaired_l2p}
\begin{center}
\resizebox{0.88\columnwidth}{!}{
\begin{tabular}{ccccc}
 \toprule 
\shortstack{Dataset\\ { }} & \shortstack{CycleGAN\\ { }} & \shortstack{C-ASL\\(Multi-class)} & \shortstack{C-ASL\\(Binary)} & \shortstack{C-ASL\\(Combined)} \\
 \midrule 
Bedroom         & 84.2 & 86.9 & 85.6 & \textbf{78.9}       \\
Cityscapes         & 83.0 & 70.9 & \textbf{64.8} & 66.0          \\
 \bottomrule 
\end{tabular}}
\end{center}
\vspace{-5mm}
\end{table}

\begin{table}
\caption{Comparison with Pix2Pix~\cite{pix2pix} on \textbf{paired label-to-photo translation} task in terms of \textbf{FID scores}, lower is better.}
\label{tab:advsegloss_fid_scores_paired_l2p}
\begin{center}
\resizebox{0.88\columnwidth}{!}{
\begin{tabular}{ccccc}
 \toprule 
\shortstack{Dataset\\ { }} & \shortstack{Pix2Pix\\ { }} & \shortstack{P-ASL\\(Multi-class)} & \shortstack{P-ASL\\(Binary)} & \shortstack{P-ASL\\(Combined)} \\
 \midrule 
Bedroom         & 128.1 & 118.2 & 122.3  & \textbf{110.1}     \\
Cityscapes      & 79.5 & 78.4  & \textbf{72.9} & 77.6          \\
 \bottomrule 
\end{tabular}}
\end{center}
\end{table}

\subsection{Label to Photo Translation}

We also experimented with label-to-photo (L2P) translation task to show the effectiveness of our model in a different task where adversarial segmentation loss could be helpful. 
In L2P task, we use ADE20k bedroom and Cityscapes datasets. Similar to S2P task, all images are resized to $256x256$ pixels. We train L2P models for $200$ epochs using the Adam optimizer~\cite{adam} with a learning rate of $0.0002$. 
We show FID scores for unpaired L2P in Table~\ref{tab:advsegloss_fid_scores_unpaired_l2p} and paired translation in Table~\ref{tab:advsegloss_fid_scores_paired_l2p}. For unpaired translation our best performing method improves the baseline for more than $5$ points on Bedroom and almost $20$ points on Cityscapes datasets. Similarly on the paired translation, the improvements regard to the baseline reaches $18$ points.

In Table~\ref{tab:advsegloss_miou_scores_unpaired_l2p} and Table~\ref{tab:advsegloss_miou_scores_paired_l2p}, we present mIoU scores for unpaired and paired L2P translation, respectively. For both cases, our best performing variant outperforms the baseline method.

We present visual results in Figure~\ref{fig:city-results} for only the baseline model and our best performing setting \textit{Binary} for unpaired translation. Our model generates more photo-realistic images, also generated images comply with the input label maps better.

\begin{table}
\caption{Comparison with CycleGAN on \textbf{unpaired label-to-photo translation}  task in terms of \textbf{mIoU scores}, higher is better.}
\label{tab:advsegloss_miou_scores_unpaired_l2p}
\begin{center}
\resizebox{\columnwidth}{!}{
\begin{tabular}{cccccc}
 \toprule 
\shortstack{Dataset\\ { }} & \shortstack{CycleGAN\\ { }}  & \shortstack{C-ASL\\(Multi-class)} & \shortstack{C-ASL\\(Binary)} & \shortstack{C-ASL\\(Combined)} & \shortstack{Oracle \\ { }}\\
 \midrule 
Bedroom         & 5.70 & 5.88 & 6.10 & \textbf{6.69} &  20.62    \\
Cityscapes      & 20.13 & \textbf{21.45} & 20.70 & 19.61  &  44.85      \\
 \bottomrule 
\end{tabular}}
\end{center}
\vspace{-5mm}
\end{table}

\begin{table}
\caption{Comparison with Pix2Pix on \textbf{paired label-to-photo translation} task in terms of \textbf{mIoU scores}, higher is better.}
\begin{center}
\resizebox{\columnwidth}{!}{
\begin{tabular}{cccccc}
 \toprule 
\shortstack{Dataset\\ { }}  &
\shortstack{Pix2Pix\\ { }}  & \shortstack{P-ASL\\(Multi-class)} & \shortstack{P-ASL\\(Binary)} & \shortstack{P-ASL\\(Combined)} & \shortstack{Oracle \\ { }}\\
 \midrule 
Bedroom         & 1.56 & 1.59 &  1.54 & \textbf{1.62} &  20.62    \\
Cityscapes      & 8.62 & 8.66  & \textbf{8.69} & 8.56  &  44.85      \\
 \bottomrule 
\end{tabular}}
\end{center}
\label{tab:advsegloss_miou_scores_paired_l2p}
\vspace{-5mm}
\end{table}

\begin{figure}[h]
\captionsetup[subfigure]{labelformat=empty}
\centering
\begin{subfigure}[b]{0.10\textwidth}
\caption{Input}
\includegraphics[width=1.0\textwidth]{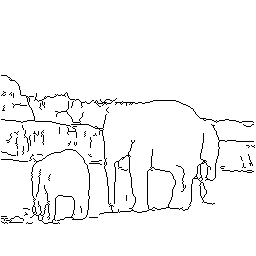}
\vspace{0.05 cm}
\includegraphics[width=1.0\textwidth]{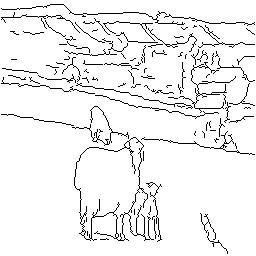}
\end{subfigure}
\begin{subfigure}[b]{0.10\textwidth}
\caption{GT}
\includegraphics[width=1.0\textwidth]{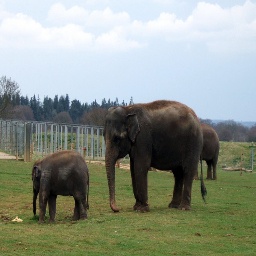}
\vspace{0.05 cm}
\includegraphics[width=1.0\textwidth]{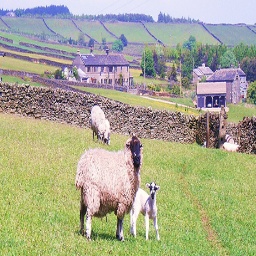}
\end{subfigure}
\begin{subfigure}[b]{0.10\textwidth}
\caption{CycleGAN}
\includegraphics[width=1.0\textwidth]{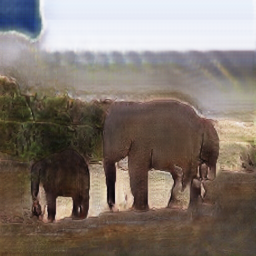}
\vspace{0.05 cm}
\includegraphics[width=1.0\textwidth]{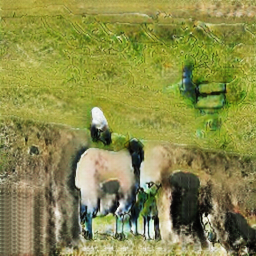}
\end{subfigure}
\begin{subfigure}[b]{0.10\textwidth}
\caption{+ASL}
\includegraphics[width=1.0\textwidth]{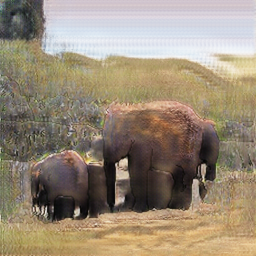}
\vspace{0.05 cm}
\includegraphics[width=1.0\textwidth]{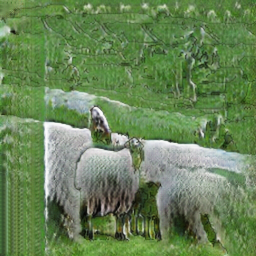}
\end{subfigure}
\caption{Sample results on elephant and sheep datasets for CycleGAN and our best model. Realism of both models are not satisfactory, however, especially colors of BG areas are better in our results. }
\label{fig:realism-res}
\vspace{-5mm}
\end{figure}

\subsection{Limitations}
\label{sec:advsegloss_limitations}

Figure~\ref{fig:realism-res} presents examples on elephant and sheep datasets where both baseline and our best performing model suffer from low visual realism. The main reason for that is these datasets contains complex foreground and background objects. However, our method performs significantly better than the baseline. Especially on the first row, colorized image using our method resembles more to the ground truth image.  

\section{Conclusion}
\label{sec:conclusion}

In this study, we present a new method for the sketch colorization problem. Our method utilizes a general purpose image segmentation network and adds an adversarial segmentation loss (ASL) to the regular GAN loss. ASL could be integrated to any GAN model, and works even if the dataset does not have segmentation labels. We used CycleGAN and Pix2Pix as baseline GAN models.
We conducted extensive evaluations on various datasets including bedroom, sheep, elephant and illustration images and evaluate the performance both quantitatively (using FID and mIoU scores) and qualitatively (through a user study). We showed that our model outperforms baselines on all datasets on both FID score and user study analysis.

Regarding the limitations of our method, although we improve the baseline both qualitatively and quantitatively, especially elephant and sheep results lack realism. Even the paired training results are not visually appealing on these two datasets, most probably due to the fact that the baseline models are not very successful at generating complex scenes. 

\section*{Acknowledgment}

The numerical calculations reported in this paper were fully performed at TUBITAK ULAKBIM, High Performance and Grid Computing Center (TRUBA resources). Dr. Akbas is supported by the ``Young Scientist Awards Program (BAGEP)'' of Science Academy, Turkey.



\bibliographystyle{elsarticle-num}
\bibliography{refs}

\end{document}